\documentclass[conference]{IEEEtran}

\usepackage{booktabs,ragged2e}
\usepackage{graphicx}
\usepackage[flushleft]{threeparttable}

\usepackage{xcolor}
\usepackage{cite}
\usepackage{amsmath}
\usepackage{url}
\usepackage{balance}
\usepackage{multirow}
\usepackage{array}
\usepackage{hyperref}
\usepackage{microtype}

\author{
\IEEEauthorblockN{Ashfaq Ali Shafin}
\IEEEauthorblockA{
Augustana College, USA\\
Florida International University, USA\\
Email: shafinashfaqali21@gmail.com
}
\and
\IEEEauthorblockN{Khandaker Mamun Ahmed}
\IEEEauthorblockA{
Dakota State University, USA\\
Email: khandakermamun.ahmed@dsu.edu
}
}

\begin{document}

\title{Leakage-Safe and Scheduler-Aware Machine Learning for Grid Job Runtime Prediction}

\maketitle

\begin{abstract}
Accurate job runtime prediction can improve scheduling-aware resource management in grid and distributed computing environments, but prediction models must be evaluated under realistic deployment constraints. This paper revisits CPU burst time prediction on the GWA-T-4 AuverGrid workload trace and reformulates it as leakage-safe pre-execution job runtime prediction. We define the target as job-level runtime, use only submission-time attributes, exclude post-execution variables, and evaluate models under temporal and cold-start settings rather than relying only on random cross-validation. We compare standard regressors, chronological historical baselines, categorical encoding strategies, and CatBoost with native categorical handling. We further add temporal hyperparameter tuning, runtime predictability analysis, feature ablation, error analysis by job length, and a minimal scheduling simulation. After temporal-validation tuning, CatBoost achieves the strongest deployment-oriented result with $R^2=0.239$, MAE $=27{,}019$, RMSE $=46{,}587$, and LogMAE $=2.646$ on the held-out temporal test set. A single-server simulation over all 69,523 held-out temporal test jobs shows that prediction-informed SJF reduces average waiting time by 50.92\% relative to FCFS. The results show that random-split evaluation overestimates performance, categorical-native boosting improves temporal generalization, and long-job underestimation remains a scheduler-relevant challenge.
\end{abstract}



\section{Introduction}
Process and job scheduling are fundamental functions in operating systems, grid platforms, and distributed computing environments. In a multitasking operating system, a process scheduler determines which process should be assigned to the CPU. Common scheduling algorithms include First Come First Serve (FCFS), Round-Robin (RR), Priority Scheduling, Shortest Job First (SJF), and Shortest Remaining Time First (SRTF) \cite{silber}. In grid and high-performance computing environments, scheduling decisions affect waiting time, turnaround time, slowdown, fairness, resource utilization, and throughput.

A CPU burst refers to the interval during which a process executes on the CPU before it blocks, terminates, or leaves the running state \cite{amur}. In production grid workload traces, however, the available target is usually job-level runtime rather than fine-grained CPU bursts between I/O operations. Therefore, this paper uses the term \emph{CPU burst time prediction} in the scheduling sense, but defines the measurable prediction target precisely as the job-level runtime recorded in the GWA-T-4 AuverGrid trace. This definition avoids ambiguity between operating-system-level CPU bursts and grid-workload job durations.

Reliable runtime estimates are useful for multiple scheduling policies. SJF and SRTF prioritize jobs with shorter expected execution times. Backfilling and reservation-based schedulers use runtime estimates to decide whether a job can be executed without delaying a reserved higher-priority job. Adaptive variants of Round-Robin scheduling may also use runtime distributions to determine time quanta. In practice, however, the actual runtime of a newly submitted job is unknown before execution. A scheduler must estimate runtime using only information available at or before job submission.

This requirement creates a methodological challenge. Workload traces contain both pre-execution variables and post-execution outcomes. If post-execution variables such as observed runtime, waiting time, job status, used memory, CPU time used, partition assignment, or last run site are used as input features, the prediction task becomes invalid for pre-scheduling deployment. Such variables are unavailable when the scheduler must make its decision and can cause target leakage. A model may then appear highly accurate because it indirectly observes the target or future execution outcomes.

Machine learning offers a promising approach for runtime prediction because supervised models can learn relationships among job submission attributes, requested resources, user behavior, group membership, queue assignment, site information, and executable identifiers. Prior work has applied learning-based approaches to runtime prediction and scheduling-related decisions \cite{MLOSNegi,MLOSShulga,smith,Helmy}. However, for machine learning to be useful in scheduling-aware systems, evaluation must be leakage-safe, time-aware, and connected to scheduling behavior rather than reported only as standalone regression accuracy.

This paper revises and strengthens CPU burst time prediction using the GWA-T-4 AuverGrid workload trace \cite{gwa-t-4-auvergrid}. The central contribution is not a new regression algorithm, but a rigorous deployment-aware reevaluation of grid runtime prediction under realistic information constraints. We define a strict pre-execution prediction task, exclude all attributes not known at submission time, fit preprocessing only on training partitions, evaluate random and deployment-oriented splits, add chronological-safe historical baselines, compare categorical encoding strategies, tune CatBoost using a temporal validation split, analyze runtime predictability, explain feature ablation results, and add a minimal scheduling simulation.

The study addresses the following research questions:

\begin{itemize}
    \item \textbf{RQ1:} How accurately can machine learning models predict grid job runtime using only attributes available at job submission time?
    \item \textbf{RQ2:} How do learning-based models compare with chronological historical baselines that use only past jobs relative to each test job?
    \item \textbf{RQ3:} How do categorical encoding and native categorical modeling affect temporal generalization?
    \item \textbf{RQ4:} Which pre-submit feature groups contribute most to temporal prediction performance?
    \item \textbf{RQ5:} How do prediction errors differ across short, medium, long, very long, and extreme-duration jobs?
    \item \textbf{RQ6:} Can prediction-informed scheduling improve waiting time in a minimal scheduling simulation?
\end{itemize}

The main contributions are:

\begin{itemize}
    \item We clarify the prediction target and define CPU burst time prediction as job-level runtime prediction on the AuverGrid trace.
    \item We introduce a leakage-safe feature policy that uses only pre-submit attributes and excludes post-execution outcomes.
    \item We evaluate chronological-safe historical baselines that prevent future information from influencing baseline predictions.
    \item We compare one-hot, ordinal, target-mean, and native categorical modeling strategies, including temporal-validation-tuned CatBoost.
    \item We add runtime predictability analysis, feature ablation, permutation importance, completed-job filtering, and error-bias analysis by job-length group.
    \item We provide a minimal scheduling simulation showing that prediction-informed SJF can reduce waiting time relative to FCFS in a simplified non-preemptive setting.
\end{itemize}

\section{Background and Related Work}

\subsection{Classical Runtime and Burst-Time Estimation}
Several scheduling algorithms require estimates of future burst time or job duration. A classical approach is exponential averaging, where the next predicted burst is computed from the most recent observed burst and the previous prediction:

\begin{equation}
\tau_{n+1}=\alpha T_n+(1-\alpha)\tau_n,
\end{equation}

where $T_n$ is the most recent observed runtime, $\tau_n$ is the previous estimate, and $\alpha$ controls the influence of recent history. This formulation is common in operating-systems scheduling discussions \cite{silber,Vandana}. Its advantage is simplicity and low computational overhead, but it uses limited information and may be insufficient for heterogeneous grid workloads.

Optimization-based and rule-based approaches have also been explored. Maheshwari et al. proposed a scheduling approach using Linear Programming Models Dual Simplex Algorithm (LPMDSA), where burst time, waiting time, and turnaround time are represented mathematically and optimized using the dual simplex method \cite{mahesh}. Pourali et al. proposed an intelligent fuzzy system to estimate CPU burst time from previous process behavior \cite{pourali}. These studies demonstrate that historical execution behavior can be useful, but they often require manually designed rules or problem-specific assumptions.

\subsection{Machine Learning for Runtime Prediction}
Machine learning has been widely explored for runtime prediction and scheduling-related decisions. Smith et al. proposed predicting parallel application runtimes by identifying similar past applications and using search methods to determine which application characteristics define similarity \cite{smith}. Negi et al. applied machine learning to CPU scheduling in Linux, aiming to reduce process turnaround time by learning CPU time-slice utilization behavior \cite{MLOSNegi}. Shulga et al. studied learned CPU/GPU execution selection in heterogeneous environments \cite{MLOSShulga}. Helmy et al. studied CPU burst time estimation on the GWA-T-4 AuverGrid dataset using machine learning \cite{Helmy}. Our work revisits this direction with stricter feature availability, chronological baselines, categorical-native modeling, temporal tuning, and scheduler-level validation.

\subsection{Scheduling Relevance and Evaluation Realism}
Recent HPC runtime-prediction work emphasizes that prediction should be connected to scheduling decisions rather than evaluated only as a standalone regression problem. Duration-Informed Workload Scheduler (DIWS) integrates machine-learning runtime estimates into an HPC scheduling policy and evaluates the scheduling effect using simulation \cite{LoretiDIWS2026}. RLBackfilling shows that better runtime prediction alone may not always yield better backfilling decisions because there is a trade-off between prediction accuracy and backfilling opportunities \cite{KolkerHicksRLBackfilling2024}. Best-practice studies for HPC runtime prediction emphasize careful methodological design and operationally meaningful evaluation \cite{MenearHPCPrediction2023}. Digital-twin-based scheduling frameworks further show that prediction and simulation can be combined to evaluate adaptive scheduling policies before deployment \cite{MaiterthHPCDigitalTwins2025,ZhangSchedTwin2025}. These works motivate our deployment-aware prediction benchmark and our minimal scheduler-level validation.

\section{Dataset and Feature Policy}

\begin{table}[t]
\centering
\caption{Dataset statistics before and after filtering.}
\label{tab:filtering}
\begin{tabular}{lcc}
\toprule
\textbf{Statistic} & \textbf{Raw Trace} & \textbf{Final Dataset} \\
\midrule
Jobs & 404,176 & 347,611 \\
Columns & 29 & 18 \\
Temporal training jobs & -- & 278,088 \\
Temporal test jobs & -- & 69,523 \\
\bottomrule
\end{tabular}
\end{table}

\begin{table*}[t]
\centering
\caption{Leakage-safe feature policy for runtime prediction.}
\label{tab:feature_policy}
\begin{tabular}{lll}
\toprule
\textbf{Attribute} & \textbf{Category} & \textbf{Use} \\
\midrule
SubmitTime & Pre-submit metadata & Used \\
SubmitTime-derived features & Temporal metadata & Used \\
ReqNProcs & Pre-submit request & Used \\
ReqTime & Pre-submit request & Used \\
ReqMemory & Pre-submit request & Used \\
UserID & Pre-submit identifier & Used with training-only encoding \\
GroupID & Pre-submit identifier & Used with training-only encoding \\
ExecutableID & Pre-submit application identifier & Used with training-only encoding \\
QueueID & Pre-submit queue identifier & Used with training-only encoding \\
OrigSiteID & Pre-submit source/site metadata & Used with training-only encoding \\
Job & Unique identifier & Excluded to prevent memorization \\
RunTime & Target/post-execution & Target only; excluded from features \\
WaitTime & Post-submission outcome & Excluded \\
NProc/NProcs & Allocation or execution outcome & Excluded unless known at submission \\
UsedCPUTime & Post-execution outcome & Excluded \\
AverageCPUTimeUsed & Post-execution outcome & Excluded \\
UsedMemory & Post-execution outcome & Excluded \\
Status & Post-execution outcome & Excluded from prediction features \\
PartitionID & Allocation outcome & Excluded \\
LastRunSiteID & Post-execution outcome & Excluded \\
\bottomrule
\end{tabular}
\end{table*}

\subsection{GWA-T-4 AuverGrid Workload Trace}
This study uses the GWA-T-4 AuverGrid workload dataset \cite{gwa-t-4-auvergrid, CETINSKI2015191} (see Table~\ref{tab:filtering}). AuverGrid is a production grid platform composed of multiple clusters located in the Auvergne region of France. The trace contains job submission and execution records, including submission time, requested resources, user and group identifiers, executable identifiers, queue information, runtime, and post-execution outcomes.

The original dataset contains 404,176 jobs and 29 columns. After filtering invalid runtime and submission-time records and applying the leakage-safe feature policy, the main experimental dataset contains 347,611 jobs.

\subsection{Target Definition}
The prediction target is the job-level runtime recorded in the AuverGrid trace. We denote this value as $y_i$ for job $i$. The trace does not provide fine-grained CPU bursts between I/O operations. Therefore, throughout the experimental sections, CPU burst time refers to the runtime that a scheduler would like to estimate before job execution.

\subsection{Leakage-Safe Feature Policy}
Table~\ref{tab:feature_policy} lists the revised feature policy. Only pre-submit fields are eligible as features. Post-execution variables are excluded because they are not available when the scheduler must make a decision.

The final feature set contains 11 numeric features and 5 categorical features. The numeric features are SubmitTime, SubmitTimeSinceStart, cyclic hour/day/week features, ReqNProcs, ReqTime, and ReqMemory. The categorical features are UserID, GroupID, ExecutableID, QueueID, and OrigSiteID.

\subsection{SubmitTime Feature Engineering}
Raw submission time can be difficult for models to use directly. We therefore derive temporal features that capture both monotonic trend and periodic workload structure. Let $s_i$ be the submission time for job $i$ and $s_0$ be the earliest submission time in the retained trace. We define:

\begin{equation}
\Delta s_i=s_i-s_0.
\end{equation}

Hour-of-day, day-of-week, and week-of-year signals are represented using sine and cosine transformations:

\begin{equation}
\sin(2\pi z/P), \quad \cos(2\pi z/P),
\end{equation}

where $z$ is the corresponding hour, day, or week index and $P$ is the period length. This representation avoids imposing artificial discontinuities between adjacent times, such as hour 23 and hour 0.

\section{Methodology}

\subsection{Prediction Task}
We formulate grid job runtime prediction as supervised regression. For each job $i$, the input vector $\mathbf{x}_i$ contains only pre-submit attributes, and the target $y_i$ is observed runtime. The goal is to learn $f(\mathbf{x}_i)$ that predicts $\hat{y}_i$ before the job executes.

\subsection{Models and Encoding Strategies}
We evaluate linear, distance-based, tree-based, and boosting models. The model set includes Linear Regression, Linear SVR, K-NN, Decision Tree, Random Forest, Gradient Boosting, XGBoost, and CatBoost. Because categorical identifiers are central to workload prediction, we compare one-hot encoding, ordinal encoding, target-mean encoding, and CatBoost's native categorical handling. K-NN is evaluated with one-hot encoding because ordinal codes impose artificial distances between unordered identifiers.

CatBoost is included because it natively handles categorical variables using ordered target statistics and can reduce the need for high-dimensional one-hot encodings. This is particularly relevant for workload traces where user and executable identifiers carry historical behavioral information.

\begin{table}[t]
\centering
\caption{Runtime distribution and predictability analysis. Entropy is estimated on $\log(1+\text{RunTime})$.}
\label{tab:predictability}
\begin{tabular}{lr}
\toprule
\textbf{Statistic} & \textbf{Value} \\
\midrule
Jobs with positive runtime & 339,314 \\
Mean runtime & 25,802 \\
Median runtime & 2,658 \\
Runtime standard deviation & 41,083 \\
Maximum runtime & 1,575,814 \\
95th percentile & 97,796 \\
99th percentile & 173,200 \\
Log-runtime entropy & 5.980 bits \\
Normalized log-runtime entropy & 0.921 \\
\bottomrule
\end{tabular}
\end{table}

\subsection{Chronological Historical Baselines}
Historical baselines can accidentally use future information under random splits if they summarize all training records without respecting each test job's submission time. To avoid this issue, we implement chronological-safe baselines. For a test job submitted at time $s_i$, user, group, and executable baselines use only training jobs from the same entity with submission time less than $s_i$. If no such history exists, the predictor falls back to the global training median. This is applied to median and exponential-moving-average baselines.

\subsection{Evaluation Protocol}
The revised evaluation includes random, temporal, user cold-start, and executable cold-start splits. The main deployment-oriented split is temporal 80/20: the earliest 80\% of jobs are used for training and the latest 20\% for testing. Hyperparameter tuning uses an inner temporal validation split inside the training period and never uses the test partition for model selection.

\subsection{CatBoost Temporal Tuning}
To ensure reproducibility, we tuned the CatBoost model using the inner temporal validation split. The hyperparameter search space included tree depths of ${4,6,8}$, iteration counts of ${200,300,500}$, learning rates of ${0.03,0.05,0.10}$, and $L_2$ leaf regularization values of ${1,3,5}$. The selected configuration used a tree depth of 8,500 iterations, a learning rate of 0.03, and an $L_2$ leaf regularization value of 5. The model was subsequently refit using this configuration on the full temporal training set and evaluated once on the held-out temporal test set.

\subsection{Runtime Predictability Analysis}
To contextualize low temporal $R^2$ values, we analyze the intrinsic predictability of the workload. Because runtime is long-tailed, we estimate histogram entropy on $\log(1+\text{RunTime})$. We report entropy in bits and normalized entropy relative to the maximum entropy of the nonempty histogram bins. This analysis does not prove an irreducible error bound, but it provides a descriptive measure of runtime uncertainty.

\subsection{Minimal Scheduling Simulation}
To connect prediction quality to scheduling behavior, we implement a minimal single-server non-preemptive scheduling simulation on the full held-out temporal test set. The simulation contains all 69,523 jobs from the latest 20\% of the trace and compares three policies: FCFS, prediction-informed SJF using tuned CatBoost runtime estimates, and oracle SJF using true runtimes. Oracle SJF is not deployable and is included only as an upper bound. The simulation is intentionally simplified and should not be interpreted as a full HPC backfilling simulator. Its purpose is to provide a scheduler-level evaluation showing that prediction errors can affect waiting time and turnaround time under the complete held-out temporal workload.
\subsection{Metrics}
We evaluate prediction performance using the coefficient of determination ($R^2$), mean absolute error (MAE), and root mean squared error (RMSE). Because grid runtimes exhibit a long-tailed distribution, we additionally report the mean absolute error on the log-transformed runtimes (Log-MAE). The corresponding equations for these evaluation metrics are provided below.

\begin{equation}
R^2(y,\hat{y}) = 1 - \frac{\sum_{i=1}^{n}(y_i-\hat{y}_i)^2}{\sum_{i=1}^{n}(y_i-\bar{y})^2}
\end{equation}

\begin{equation}
MAE=\frac{1}{n}\sum_{i=1}^{n}|y_i-\hat{y}_i|
\end{equation}

\begin{equation}
RMSE=\sqrt{\frac{1}{n}\sum_{i=1}^{n}(y_i-\hat{y}_i)^2}
\end{equation}

\begin{equation}
LogMAE=\frac{1}{n}\sum_{i=1}^{n}|\log(1+y_i)-\log(1+\hat{y}_i)|.
\end{equation}

\begin{table}[t]
\centering
\caption{Tuned CatBoost performance on the temporal 80/20 split.}
\label{tab:catboost_tuned}
\begin{tabular}{lrrrr}
\toprule
\textbf{Model} & \textbf{$R^2$} & \textbf{MAE} & \textbf{RMSE} & \textbf{LogMAE} \\
\midrule
CatBoost & 0.239 & 27,019 & 46,587 & 2.646 \\
\bottomrule
\end{tabular}
\end{table}

For scheduling simulation, we report average waiting time, median waiting time, average turnaround time, and percentage improvement in average waiting time relative to FCFS.

\section{Experimental Results}

\subsection{Runtime Distribution and Predictability}
Table~\ref{tab:predictability} summarizes the runtime distribution. The retained workload is highly skewed: the median runtime is 2,658 seconds, while the mean is 25,802 seconds and the maximum observed runtime is 1,575,814 seconds. The 95th and 99th percentiles are 97,796 and 173,200 seconds, respectively. The normalized entropy of $\log(1+\text{RunTime})$ is 0.921, indicating substantial uncertainty even after log transformation. This supports interpreting modest temporal $R^2$ values in the context of a highly long-tailed and nonstationary production workload.

\begin{table*}[t]
\centering
\caption{Best results across evaluation settings. Random splitting is optimistic; temporal and cold-start splits better reflect deployment.}
\label{tab:main_summary}
\begin{tabular}{llrrrr}
\toprule
\textbf{Setting} & \textbf{Best Model} & \textbf{$R^2$} & \textbf{MAE} & \textbf{RMSE} & \textbf{LogMAE} \\
\midrule
Leakage-safe random 5-fold & K-NN & 0.700 & 7,417 & 22,332 & 0.679 \\
Temporal 80/20, conventional models & Linear Regression & 0.085 & 33,589 & 51,095 & 2.777 \\
Temporal 80/20, native categorical & CatBoost & \textbf{0.239} & 27,019 & 46,587 & 2.646 \\
User cold-start & Gradient Boosting & 0.133 & 23,747 & 36,988 & 3.425 \\
Executable cold-start & K-NN & 0.207 & 17,680 & 38,072 & 2.126 \\
\bottomrule
\end{tabular}
\end{table*}

\begin{table*}[t]
\centering
\caption{Chronological-safe historical baselines. For each test job, entity history uses only training jobs submitted earlier than the test job.}
\label{tab:chrono_baselines}
\begin{tabular}{llrrrr}
\toprule
\textbf{Split} & \textbf{Baseline} & \textbf{$R^2$} & \textbf{MAE} & \textbf{RMSE} & \textbf{LogMAE} \\
\midrule
Random 80/20 & Global Median & -0.311 & 24,828 & 46,820 & 3.111 \\
Random 80/20 & User Median & 0.131 & 18,391 & 38,105 & 1.851 \\
Random 80/20 & Group Median & -0.177 & 23,489 & 44,360 & 2.651 \\
Random 80/20 & Executable Median & -0.315 & 24,778 & 46,881 & 3.091 \\
Random 80/20 & User EMA & \textbf{0.638} & \textbf{8,847} & \textbf{24,587} & \textbf{0.949} \\
Random 80/20 & Executable EMA & 0.305 & 17,412 & 34,092 & 2.061 \\
\midrule
Temporal 80/20 & Global Median & -0.300 & 31,031 & 60,905 & 2.740 \\
Temporal 80/20 & User Median & \textbf{0.064} & \textbf{27,016} & \textbf{51,682} & 2.371 \\
Temporal 80/20 & Group Median & -0.078 & 29,607 & 55,457 & \textbf{2.241} \\
Temporal 80/20 & Executable Median & -0.298 & 30,952 & 60,865 & 2.711 \\
Temporal 80/20 & User EMA & -0.157 & 31,659 & 57,455 & 2.262 \\
Temporal 80/20 & Executable EMA & -0.031 & 33,702 & 54,226 & 2.923 \\
\bottomrule
\end{tabular}
\end{table*}

\begin{table*}[t]
\centering
\caption{Temporal 80/20 performance under categorical encoding strategies. CatBoost uses native categorical handling and temporal-validation-selected hyperparameters.}
\label{tab:encoding}
\begin{tabular}{llrrrr}
\toprule
\textbf{Model} & \textbf{Encoding} & \textbf{$R^2$} & \textbf{MAE} & \textbf{RMSE} & \textbf{LogMAE} \\
\midrule
Linear Regression & One-hot & -0.171 & 43,524 & 57,797 & 3.141 \\
Random Forest & One-hot & -0.005 & 29,184 & 53,536 & 2.479 \\
Gradient Boosting & One-hot & 0.062 & 30,131 & 51,723 & 2.675 \\
XGBoost & One-hot & -0.023 & 31,191 & 54,016 & 2.563 \\
Linear Regression & Ordinal & -0.122 & 45,237 & 56,573 & 3.254 \\
Random Forest & Ordinal & -0.051 & 33,478 & 54,759 & 2.709 \\
Gradient Boosting & Ordinal & 0.097 & 30,583 & 50,770 & 2.650 \\
XGBoost & Ordinal & -0.010 & 31,938 & 53,677 & 2.723 \\
Linear Regression & Target mean & 0.039 & 38,234 & 52,374 & 3.054 \\
Random Forest & Target mean & 0.030 & 32,799 & 52,599 & 2.502 \\
Gradient Boosting & Target mean & 0.107 & 28,299 & 50,467 & 3.182 \\
XGBoost & Target mean & 0.082 & 30,888 & 51,177 & 2.695 \\
CatBoost & Native categorical & \textbf{0.239} & \textbf{27,019} & \textbf{46,587} & 2.646 \\
\bottomrule
\end{tabular}
\end{table*}

\subsection{Main Temporal Result}
Table~\ref{tab:catboost_tuned} reports the tuned CatBoost result on the temporal 80/20 split. Hyperparameters are selected using only the inner temporal validation split. The tuned CatBoost model achieves $R^2=0.239$, MAE $=27{,}019$, RMSE $=46{,}587$, and LogMAE $=2.646$ on 69,523 held-out later jobs. The model trains in 17.18 seconds and predicts the test set in 0.10 seconds.

\begin{table*}[t]
\centering
\caption{CatBoost feature ablation on the temporal 80/20 split.}
\label{tab:catboost_ablation}
\begin{tabular}{lrrrr}
\toprule
\textbf{Feature Group} & \textbf{$R^2$} & \textbf{MAE} & \textbf{RMSE} & \textbf{LogMAE} \\
\midrule
All features & \textbf{0.239} & \textbf{27,019} & \textbf{46,587} & 2.646 \\
Identifiers + requested resources & 0.129 & 29,149 & 49,850 & 2.401 \\
Identifiers only & 0.126 & 29,293 & 49,945 & \textbf{2.333} \\
Raw submit-time only & -0.009 & 35,068 & 53,667 & 3.000 \\
Cyclic submit-time only & -0.027 & 36,710 & 54,129 & 3.033 \\
Requested resources only & -0.031 & 33,454 & 54,232 & 2.926 \\
All submit-time features & -0.033 & 37,734 & 54,289 & 3.070 \\
Requested resources + time & -0.045 & 34,298 & 54,600 & 2.945 \\
\bottomrule
\end{tabular}
\end{table*}

\subsection{Random, Temporal, and Cold-Start Summary}
Table~\ref{tab:main_summary} summarizes the strongest results across evaluation settings. Random splitting gives the strongest apparent performance, while temporal and cold-start settings are substantially more difficult. The tuned CatBoost temporal result is lower than the earlier untuned CatBoost value but is more reproducible because hyperparameters are selected without using the held-out temporal test set.

\subsection{Chronological-Safe Historical Baselines}
Table~\ref{tab:chrono_baselines} reports chronological-safe historical baselines. Under random 80/20 splitting, user-level exponential moving average performs strongly with $R^2=0.638$ and MAE $=8{,}847$. This shows that user history is highly informative when similar user behavior is present in both training and test data. Under temporal splitting, however, chronological baselines are weaker. User median achieves $R^2=0.064$, while several other entity-level baselines have negative $R^2$ values. These results confirm that chronological consistency matters and that historical repetition alone does not fully solve future-period generalization.

\subsection{Categorical Encoding and CatBoost}
Table~\ref{tab:encoding} compares categorical encoding strategies on the temporal split. Target-mean encoding improves several conventional models compared with one-hot and ordinal encoding. Gradient Boosting with target encoding obtains the best conventional encoding-comparison result with $R^2=0.107$. Tuned CatBoost with native categorical handling performs substantially better, reaching $R^2=0.239$.

This result supports the use of categorical-native boosting for workload traces with high-value user, group, and executable identifiers. It also shows that ordinal encoding can be misleading because it imposes arbitrary numeric order on unordered identifiers.

\subsection{Feature Ablation}
The earlier Random Forest ablation produced the counterintuitive result that the all-feature configuration performed worse than the identifier-only configuration. To examine whether this behavior was model-specific, Table~\ref{tab:catboost_ablation} repeats the ablation using the tuned CatBoost model. The all-feature configuration achieves the strongest performance with $R^2=0.239$, compared with $R^2=0.126$ for the identifier-only configuration, while the requested-resource-only configuration yields a negative $R^2$. These results indicate that the earlier Random Forest degradation does not imply that the additional features are uninformative. Instead, it likely reflects the difficulty of modeling sparse, high-cardinality identifiers and their interactions with temporal and resource-related features using one-hot-encoded tree ensembles. CatBoost's native categorical handling more effectively integrates identifier, requested-resource, and temporal information.

\subsection{Completed-Only Label Quality Analysis}
The trace contains 336,085 completed jobs, representing 96.68\% of the filtered dataset. Status values 0 and 5 account for the remaining 3.32\%. Table~\ref{tab:completed} evaluates temporal prediction using only completed jobs. The results remain low, although Gradient Boosting achieves the best completed-only $R^2=0.065$. This indicates that label-quality filtering alone does not explain the temporal generalization challenge.

\begin{table}[t]
\centering
\caption{Completed-only temporal prediction results.}
\label{tab:completed}
\begin{tabular}{lrrrr}
\toprule
\textbf{Model} & \textbf{$R^2$} & \textbf{MAE} & \textbf{RMSE} & \textbf{LogMAE} \\
\midrule
Linear Regression & -0.057 & 36,190 & 48,719 & 3.071 \\
Random Forest & 0.021 & \textbf{27,243} & 46,900 & \textbf{2.534} \\
Gradient Boosting & \textbf{0.065} & 27,735 & \textbf{45,829} & 2.736 \\
\bottomrule
\end{tabular}
\end{table}

\subsection{Permutation Importance}
Permutation importance for the temporal Random Forest model identifies UserID, GroupID, and ReqTime as the most influential features. Table~\ref{tab:importance} reports the highest-importance features. This confirms that repeated user and group behavior remains central to prediction, while requested time contributes additional signal.

\begin{table}[t]
\centering
\caption{Permutation importance for Random Forest on temporal split.}
\label{tab:importance}
\begin{tabular}{lrr}
\toprule
\textbf{Feature} & \textbf{Importance Mean} & \textbf{Std.} \\
\midrule
UserID & 0.1102 & 0.0020 \\
GroupID & 0.0605 & 0.0028 \\
ReqTime & 0.0229 & 0.0008 \\
SubmitDayCos & 0.0066 & 0.0017 \\
SubmitWeekCos & 0.0016 & 0.0003 \\
OrigSiteID & 0.0010 & 0.0005 \\
QueueID & 0.0009 & 0.0003 \\
\bottomrule
\end{tabular}
\end{table}

\subsection{Error Analysis by Job Length}
Table~\ref{tab:catboost_error_bins} reports error bias for tuned CatBoost across job-length groups. The model overestimates short and medium jobs but underestimates very long and extreme jobs. For extreme jobs, the underestimation rate is 99.97\%, with a mean error of $-105{,}448$ seconds. This is important for scheduling because underestimating long jobs can be more harmful than overestimating short jobs, especially in reservation-based and backfilling schedulers.

\begin{table*}[t]
\centering
\caption{Tuned CatBoost error analysis by job-length group. Positive error indicates overestimation; negative error indicates underestimation.}
\label{tab:catboost_error_bins}
\begin{tabular}{lrrrrr}
\toprule
\textbf{Job Group} & \textbf{N} & \textbf{Mean True} & \textbf{Mean Pred.} & \textbf{Mean Error} & \textbf{Underest. Rate} \\
\midrule
Short & 17,385 & 70 & 22,243 & 22,173 & 0.211 \\
Medium & 17,377 & 880 & 22,334 & 21,455 & 0.133 \\
Long & 17,380 & 19,595 & 23,941 & 4,346 & 0.461 \\
Very long & 13,904 & 86,300 & 54,396 & -31,903 & 0.826 \\
Extreme & 3,477 & 182,032 & 76,585 & -105,448 & 1.000 \\
\bottomrule
\end{tabular}
\end{table*}

\subsection{Minimal Scheduling Simulation}
Table~\ref{tab:minimal_scheduling} reports the minimal single-server scheduling simulation on all 69,523 held-out temporal test jobs. Predicted SJF uses tuned CatBoost runtime estimates. Oracle SJF uses true runtimes and serves only as an upper bound. Predicted SJF reduces average waiting time by 50.92\% relative to FCFS, while oracle SJF reduces average waiting time by 75.83\%.

\begin{table*}[t]
\centering
\caption{Minimal single-server scheduling simulation on all 69,523 held-out temporal test jobs. Predicted SJF uses tuned CatBoost runtime estimates. Oracle SJF uses true runtimes and serves only as an upper bound.}
\label{tab:minimal_scheduling}
\begin{tabular}{lrrrr}
\toprule
\textbf{Scheduler} & \textbf{Avg. Waiting} & \textbf{Median Waiting} & \textbf{Avg. Turnaround} & \textbf{Waiting Improvement} \\
\midrule
FCFS & 82,926,590 & 84,516,105 & 82,958,498 & 0.00\% \\
Predicted SJF & 40,698,550 & 26,872,289 & 40,730,460 & 50.92\% \\
Oracle SJF & 20,043,860 & 16,415 & 20,075,770 & 75.83\% \\
\bottomrule
\end{tabular}
\end{table*}

This simulation does not replace a full production backfilling simulator. It ignores multi-node placement, queue priorities, reservations, resource fragmentation, and backfilling constraints. However, because it is run on the complete held-out temporal test set, it provides stronger evidence than a sampled sanity check: even moderate temporal prediction accuracy can produce meaningful waiting-time improvements in a simplified non-preemptive setting. The large remaining gap between predicted SJF and oracle SJF also shows that prediction quality and long-job underestimation remain important practical limitations.

\section{Discussion}

\subsection{What the Revised Results Show}
The revised results change the interpretation of the study. High random-split scores should not be treated as deployment-valid evidence. Random splitting benefits from repeated users, groups, and applications appearing in both training and testing. Temporal and cold-start settings are more difficult and reveal the real generalization challenge.

The tuned CatBoost result shows that categorical-native boosting can recover meaningful temporal performance from pre-submit features, reaching $R^2=0.239$ on the held-out temporal period. The runtime entropy analysis helps explain why temporal performance remains modest: the workload is highly long-tailed and uncertain, with normalized log-runtime entropy 0.921. Therefore, the paper should not claim that runtime prediction is solved. Instead, it should claim that leakage-safe, categorical-native modeling provides a more realistic and reproducible benchmark for deployment-oriented prediction.

\subsection{Why CatBoost Helps}
Workload traces contain repeated behavioral patterns at the user, group, and executable levels. One-hot encoding can represent these categories but creates high-dimensional sparse features. Ordinal encoding imposes arbitrary order and can distort model behavior. Target encoding can be useful but must be applied carefully to avoid leakage. CatBoost provides an effective compromise because it handles categorical variables natively and uses ordered statistics. The ablation results show that CatBoost benefits from combining identifiers, requested resources, and temporal features, unlike the earlier Random Forest one-hot ablation.

\subsection{Scheduling Implications}
Prediction metrics do not automatically imply scheduling improvement. Schedulers may be more sensitive to underestimation than overestimation because underestimating a long job can violate a backfilling reservation. The minimal scheduling simulation over all held-out temporal test jobs provides evidence that prediction-informed scheduling can reduce waiting time relative to FCFS in a simplified single-server setting, but it also confirms that a full scheduler-integrated evaluation is needed. The next step is to evaluate prediction-informed EASY backfilling, conservative backfilling, SJF, and hybrid policies under realistic resource constraints.

\subsection{Long-Job Underestimation}
The error analysis shows that tuned CatBoost systematically underestimates very long and extreme jobs. This failure mode is scheduler-relevant. Underestimation can increase reservation violations and delay other jobs if a scheduler assumes that a long job will finish earlier than it actually does. This motivates asymmetric loss functions, quantile prediction, conformal prediction intervals, and conservative runtime estimates for backfilling.

\section{Threats to Validity}
Several threats to validity remain. First, the study uses one grid workload trace, so results may not generalize to all grid, cloud, or HPC environments. Second, the trace provides job-level runtime rather than fine-grained CPU bursts between I/O operations. Third, categorical identifiers are useful when histories exist but create cold-start challenges. Fourth, temporal validation does not fully eliminate distribution shift, as validation and test periods may differ. Fifth, although the scheduling simulation uses the complete held-out temporal test set, it is intentionally simplified and does not model a full production HPC scheduler, resource fragmentation, priorities, reservations, or EASY backfilling. Sixth, entropy analysis describes distributional uncertainty but does not provide a formal irreducible-error bound.

\section{Future Work}
Future work should integrate the predictors into a full scheduling simulator and measure waiting time, turnaround time, bounded slowdown, utilization, fairness, and reservation violations. We will evaluate prediction-informed SJF, EASY backfilling, conservative backfilling, and hybrid policies. We will also explore quantile regression, conformal prediction intervals, asymmetric loss functions that penalize long-job underestimation, online learning for user and executable histories, and cross-trace evaluation on additional workload traces. A digital-twin-style evaluation framework could further support repeated what-if comparisons of scheduling policies before deployment.

\section{Conclusion}
This paper reformulated CPU burst time prediction for grid workload scheduling as leakage-safe pre-execution job runtime prediction. The revised task defines the target as job-level runtime in the AuverGrid trace and excludes all post-execution variables from the feature set. The paper adds chronological-safe baselines, temporal feature engineering, categorical encoding comparisons, temporal-validation-tuned CatBoost, runtime predictability analysis, completed-job filtering, feature ablation, permutation importance, error-bias analysis, and a minimal scheduling simulation. The results show that random splitting substantially overestimates deployment performance. Tuned CatBoost improves temporal prediction under realistic feature constraints, while long-job underestimation remains a major scheduling concern. A minimal scheduling simulation over the complete held-out temporal test set shows that prediction-informed SJF can reduce average waiting time relative to FCFS, but full backfilling simulation remains future work. The main contribution is therefore a rigorous benchmark and diagnostic analysis showing why leakage-safe, time-aware, and scheduler-aware evaluation is necessary for grid runtime prediction.

\balance
\bibliographystyle{IEEEtran}
\bibliography{Reference}

\end{document}